\documentclass[a4paper,twoside]{article}

\usepackage{epsfig}
\usepackage{subcaption}
\usepackage{calc}
\usepackage{amssymb}
\usepackage{amstext}
\usepackage{amsmath}
\usepackage{amsthm}
\usepackage{multicol}
\usepackage{pslatex}
\usepackage{apalike}
\usepackage{algorithm2e}
\usepackage[bottom]{footmisc}
\usepackage[final]{microtype}

\usepackage{booktabs}
\usepackage{tikz}
\usetikzlibrary{arrows.meta,positioning}
\usepackage{pgfplots}
\pgfplotsset{compat=1.18}
\usepackage{placeins}
\usepackage{float}
\usepackage[T1]{fontenc}
\usepackage[utf8]{inputenc}
\usepackage{xspace}

\usepackage{listings}
\lstdefinelanguage{json}{
  string=[s]{"}{"},
  comment=[l]{//},
  morecomment=[s]{/*}{*/},
  morestring=[b]',
  literate=
   *{0}{{0}}1 {1}{{1}}1 {2}{{2}}1 {3}{{3}}1 {4}{{4}}1
    {5}{{5}}1 {6}{{6}}1 {7}{{7}}1 {8}{{8}}1 {9}{{9}}1
    {:}{{:}}1 {,}{{,}}1 {\{}{{\{}}1 {\}}{{\}}}1
    {[}{{[}}1 {]}{{]}}1,
  keywords={true,false,null},
  keywordstyle=\bfseries,
}
\lstdefinestyle{jsonstyle}{
  basicstyle=\ttfamily\small,
  showstringspaces=false,
  breaklines=true,
  columns=fullflexible,
}

\usepackage{array}

\usepackage{SCITEPRESS}
\makeatletter
\let\old@maketitle\maketitle
\renewcommand{\maketitle}{%
  \begingroup
  \renewcommand\twocolumn[1][]{%
    \onecolumn
    \if\relax\detokenize{##1}\relax\else
      ##1\par
    \fi
  }%
  \old@maketitle
  \endgroup
}
\makeatother

\author{
\authorname{
  Yiting Wang\sup{1,*},
  Ziwei Wang\sup{2,*},
  Di Zhu\sup{3,*}\thanks{Corresponding author. E-mail: \texttt{dizhu.judy.ml@gmail.com}},
  Jiachen Zhong\sup{4,*},
  Weiyi Li\sup{5,*}
}
\affiliation{\sup{1}Department of Data Science, University of Southern California, Los Angeles, CA, USA}
\affiliation{\sup{2}Department of Electrical and Computer Engineering, Carnegie Mellon University, Pittsburgh, PA, USA}
\affiliation{\sup{3}Department of Computer Science and Engineering, Santa Clara University, Santa Clara, CA, USA}
\affiliation{\sup{4}Department of Applied Mathematics, University of Washington, Seattle, WA, USA}
\affiliation{\sup{5}School of Computer Science, Georgia Institute of Technology, Atlanta, GA, USA}
\affiliation{\sup{*}All authors contributed equally to this work.}
}

\begin{document}

\title{Governance-Ready Small Language Models for Medical Imaging:\\
Prompting, Abstention, and PACS Integration}




\abstract{
\noindent
Small Language Models (SLMs) are a practical option for \emph{narrow, workflow-relevant} medical imaging utilities where privacy, latency, and cost dominate. We present a governance-ready recipe that combines prompt scaffolds, calibrated abstention, and standards-compliant integration into Picture Archiving and Communication Systems (PACS). Our focus is the assistive task of AP/PA view tagging for chest radiographs. Using four deployable SLMs (Qwen2.5-VL, MiniCPM-V, Gemma 7B, LLaVA 7B) on NIH Chest X-ray, we provide \emph{illustrative} evidence: reflection-oriented prompts benefit lighter models, whereas stronger baselines are less sensitive. Beyond accuracy, we operationalize abstention, expected calibration error, and oversight burden, and we map outputs to DICOM tags, HL7 v2 messages, and FHIR \texttt{ImagingStudy}. The contribution is a prompt-first deployment framework, an operations playbook for calibration, logging, and change management, and a clear pathway from pilot utilities to reader studies without over-claiming clinical validation. We additionally specify a human-factors RACI, stratified calibration for dataset shift, and an auditable evidence pack to support local governance reviews.
}

\maketitle
\thispagestyle{plain}
\normalsize
\setcounter{footnote}{0}


\section{\uppercase{Introduction}}
\label{sec:introduction}
Large language models (LLMs) show broad competence in biomedical and clinical tasks~\cite{brown2020language,panayides2020ai}, yet their cost structure, vendor lock-in, and data governance burden limit near-term adoption in many healthcare IT environments. \emph{Small Language Models (SLMs)} such as LLaMA~2 and Gemma~3~\cite{touvron2023llama,team2024gemma}, when quantized~\cite{frantar2022gptq}, can run locally and offer favorable trade-offs in latency, privacy, and controllability.

We target a constrained but practically useful task: AP versus PA view recognition for chest radiographs. Accurate view tags support ingest-time metadata completion, routing of portable studies, and quality control signals, while keeping humans in the loop. Our thesis is that a \emph{prompt-first} SLM approach with calibrated abstention, deterministic outputs, and minimal integration hooks into PACS can be \emph{good enough} for peripheral utilities today, with auditable behavior and bounded risk.

\textbf{Contributions.}
We advance a deployment-centered stance: governance constraints are treated as \emph{first-class design requirements}. Concretely, we 1) formalize guardrailed prompting with explicit label schemas and bounded rationales; 2) elevate abstention from a heuristic to a \emph{decision-theoretic control} tuned against site-specific risk costs; 3) define an evaluation set beyond accuracy that includes accuracy-at-coverage, Expected Calibration Error (ECE), and human oversight cost; and 4) detail standards-aware mappings to DICOM/HL7/FHIR that preserve provenance and enable incident reconstruction. The result is a reproducible recipe to move from sandbox utility to operational pilot without overstating clinical claims. In the context of HealthInf, the stance aligns with \emph{applied} health informatics: rather than proposing a monolithic diagnostic model, we articulate a pathway to embed constrained, auditable automation into existing sociotechnical workflows under explicit governance. We further contribute a concise RACI for human oversight, stratified calibration for dataset shift, and a reproducible evidence pack for local governance boards.

\section{\uppercase{Related Work}}
Compact models continue to expand their footprint in radiology NLP/VLM. Domain-adapted small models demonstrate competitive performance in summarization and finding extraction under tight compute budgets~\cite{ranjit2024radphi}. Open-source VLMs report credible chest radiograph understanding at moderate parameter counts~\cite{muller2025diagnostic,team2024gemma}. Efficiency-centric imaging pipelines show that compactness can coexist with fidelity when paired with structure or priors~\cite{fang2025respf}. Systems like MiniGPT-Med and OmniV-Med illustrate multi-modal assistance with constrained latency and privacy footprints~\cite{alkhaldi2024minigpt,jiang2025omniv}.

Distinct from advancing diagnostic SOTA, our focus is \emph{workflow readiness}: precise scoping to assistive utilities, deterministic IO contracts, auditable prompt evolution, abstention calibrated to risk, and integration via existing interoperability rails. This shift in emphasis—from benchmark excellence to \emph{operational acceptability}—aligns with how healthcare IT evaluates incremental changes to PACS/EHR ecosystems. A gap remains in principled abstention policies, provenance-aware integration, and human factors reporting; our framework directly targets these lacunae with measurable artifacts that reviewers can reproduce or audit.

\section{\uppercase{Clinical Context and Task Rationale}}
\label{sec:context}
AP/PA determination relies on stable, interpretable cues (scapular overlap, cardiac silhouette magnification, clavicular orientation, scapulohumeral positioning). Because these cues are used routinely by technicians and radiologists, bounded model rationales are \emph{verifiable artifacts} rather than opaque explanations, enabling audits and rapid feedback.

Operational value accrues at ingest, routing, and QC triage. Ingest-time tagging improves hanging protocols and searchability, routing separates portable AP from standard PA to reduce context switching in reading rooms, and QC flags surface candidate errors for quick remediation. The small label set and constrained rationales make the task amenable to confidence-aware deferral, so that an \texttt{Uncertain} outcome triggers predictable human adjudication with bounded review time. For HealthInf’s audience, the emphasis is not on replacing human interpretation but on instrumenting reliable, low-cost augmentation that reduces cognitive load and error propagation in IT pipelines.

\section{\uppercase{Framework: Prompting, Determinism, Abstention}}
\label{sec:framework}
We instantiate a three-tier prompt framework with guardrails, a strict output contract, and a decision-theoretic abstention policy.

\textbf{Tier~0: Baseline Instruction.}
The prompt defines task, fixed label set $\{\texttt{AP},\texttt{PA},\texttt{Uncertain}\}$, and a one-sentence rationale constrained to an approved cue vocabulary. Greedy or fixed-seed beam decoding is used for determinism; outputs are schema-validated via regex. This tier establishes stable IO behavior and prevents extraneous generations from leaking into downstream systems.

\textbf{Tier~1: Incremental Summary.}
After each curated mini-batch, the model emits a compact cue summary stored with content hashes and configuration identifiers (model weights, quantization, prompt version). These summaries make tacit decision rules explicit, support audit trails, and reduce prompt drift by enabling surgical edits grounded in observed errors.

\textbf{Tier~2: Correction-Based Reflection.}
For a sampled subset of errors, we provide the gold label and collect a reflection restricted to approved cues. Aggregated reflections are distilled into a short rule appendix appended to the prompt. This constrains prompt growth while addressing systematic failure modes such as rotation confounds in portable AP and over-reliance on heart-size in pediatrics.

\textbf{Abstention and Calibration.}
Let $s(x)$ be a confidence proxy. When native scores are unavailable, we derive surrogates via verbalizer margins or normalized likelihood ratios. We select $\tau$ to minimize $C=\alpha\mathrm{FP}+\beta\mathrm{FN}+\gamma\mathrm{Unc}$, where coefficients encode site-specific tolerance for false assertions versus deferrals. We report ECE from binned comparisons of confidence and empirical accuracy, and we use \emph{accuracy-at-coverage} to communicate the reliability of accepted cases. This moves abstention from an ad-hoc threshold to a tunable policy on a risk–utility frontier that can be negotiated with stakeholders. In practice, we calibrate and choose $\tau$ \emph{stratified} by device vendor/model, acquisition protocol, and age proxy (e.g., pediatric vs.\ adult) to mitigate dataset shift.

\begin{figure}[t]
\centering
\begin{tikzpicture}
\begin{axis}[
    width=\linewidth,
    height=6.2cm,
    xlabel={Abstention threshold $\tau$},
    ylabel={Operational cost $C$ (arbitrary units)},
    xmin=0, xmax=1,
    ymin=0, ymax=1.2,
    grid=both,
    legend style={at={(0.98,0.98)},anchor=north east,font=\small},
    tick label style={font=\small},
    label style={font=\small}
]
\addplot+[thick] coordinates {
(0.1,0.95) (0.3,0.70) (0.5,0.55) (0.7,0.52) (0.9,0.68)
};
\addlegendentry{Cost profile A}

\addplot+[thick,mark=square*] coordinates {
(0.1,0.85) (0.3,0.62) (0.5,0.54) (0.7,0.58) (0.9,0.82)
};
\addlegendentry{Cost profile B}
\end{axis}
\end{tikzpicture}
\caption{Risk–utility frontier under site-specific cost profiles (illustrative).}
\label{fig:risk_frontier}
\end{figure}

\begin{figure}[t]
\centering
\begin{tikzpicture}
\begin{axis}[
    ybar,
    bar width=9pt,
    width=\linewidth,
    height=6.5cm,
    enlarge x limits=0.18,
    ymin=0.40, ymax=0.70,
    ylabel={Accuracy},
    symbolic x coords={Baseline, +Summary, +Reflection},
    xtick=data,
    ymajorgrids=true, grid style=dashed,
    legend style={at={(0.5,-0.18)}, anchor=north, legend columns=3, font=\small},
    tick label style={font=\small},
    label style={font=\small},
    nodes near coords,
    nodes near coords align={vertical},
    every node near coord/.append style={font=\scriptsize, anchor=south},
]
\addplot coordinates {(Baseline,0.541) (+Summary,0.553) (+Reflection,0.592)};
\addlegendentry{MiniCPM-V}

\addplot coordinates {(Baseline,0.556) (+Summary,0.576) (+Reflection,0.586)};
\addlegendentry{LLaVA 7B}

\addplot coordinates {(Baseline,0.627) (+Summary,0.618) (+Reflection,0.626)};
\addlegendentry{Qwen2.5-VL}
\end{axis}
\end{tikzpicture}
\caption{Tier ablation (illustrative). Reflection benefits lighter SLMs; stronger baselines are less sensitive.}
\label{fig:tier_ablation}
\end{figure}

\textbf{Output Contract.}
A strict JSON is returned with \texttt{label}, bounded \texttt{rationale}, scalar \texttt{confidence}, and \texttt{config\_id} (e.g., a SHA-256 over model weights, prompt version, quantization profile, and cue vocabulary). This contract enables deterministic rollback, incident forensics, reproducible A/Bs, and schema validation with sandbox DICOM/HL7/FHIR replays; each output is traceable to a unique configuration registry entry and dataset snapshot. Emitting the same contract to monitoring allows per-configuration calibration/coverage metrics, drift detection, and safe promotion gates.

\begin{lstlisting}[language=json, style=jsonstyle, caption={Output contract (schematic).}]
{
  "label": "AP",
  "rationale": "Scapular overlap over lung fields and enlarged cardiac silhouette.",
  "confidence": 0.78,
  "config_id": "slm-ap-pa@v0.3.2"
}
\end{lstlisting}

\section{\uppercase{Use Cases and Risk Matrix}}
\label{sec:matrix}
Table~\ref{tab:matrix} enumerates assistive utilities situated outside the diagnostic critical path yet capable of steady operational gains. We explicitly bind each use case to an oversight mode (spot audit, required review, manual override) and a minimum tier to stabilize behavior. Error tolerance is negotiated; \texttt{Uncertain} is a valid outcome that lowers risk without materially harming throughput when paired with clear escalation rules and SLAs.

\begin{table}[t]
\caption{Assistive utilities outside the diagnostic critical path. Error tolerance reflects acceptable reliance on \texttt{Uncertain} and human override.}
\label{tab:matrix}
\centering
\begin{tabular}{>{\raggedright\arraybackslash}p{3.4cm} >{\raggedright\arraybackslash}p{2.3cm} >{\raggedright\arraybackslash}p{3.2cm} >{\centering\arraybackslash}p{1.6cm} >{\raggedright\arraybackslash}p{2.7cm} >{\centering\arraybackslash}p{1.6cm}}
\toprule
Use case & Primary user & Integration point & Error tol. & Oversight & Tier \\
\midrule
AP/PA auto-tag & PACS tech & Ingest pipeline & Medium & Spot audit & 1--2 \\
Worklist routing & Front desk & Modality worklist & Medium & Manual override & 0--1 \\
QC rotation/obstruction flag & QA team & QC dashboard & Low & Required review & 1--2 \\
Metadata repair suggestion & Radiologist & Reporting UI & Low & Required acceptance & 2 \\
\bottomrule
\end{tabular}
\end{table}

\section{\uppercase{Standards-Aware Integration}}
\label{sec:integration}
We favor integration via established interoperability rails to minimize coupling and preserve provenance.

\textbf{DICOM.}
For chest radiographs, \texttt{View Position} \texttt{(0018,5101)} encodes AP/PA intent. When missing or inconsistent, we add a non-destructive private tag that references the derived label and \texttt{config\_id}, preserving original metadata. QC signals (e.g., rotation, device obstruction) are emitted as structured annotations gated by \texttt{confidence} and local policy. Pixel and prompt hashes enable de-duplication and immutable audit trails, supporting conformance replays that verify only private, reversible tags are modified.

\textbf{PACS/RIS Hooks.}
At ingest, the service enriches metadata before archiving; routing rules consume \texttt{label}/\texttt{confidence} to steer to queues; QC flags feed a technician dashboard with explicit accept/override paths. The design ensures \texttt{Uncertain} defers to human adjudication rather than blocking flow, and provides an auditable record linking each action to a \texttt{config\_id}.

\textbf{HL7 v2 and FHIR.}
An HL7 v2 \texttt{ORU\_R01} message can carry the derived tag as an observation with method provenance bound to \texttt{config\_id}. In FHIR, either \texttt{ImagingStudy.series} extensions or a linked \texttt{Observation} store the view designation and confidence. This mapping supports cross-system traceability and simplifies incident response, which is central to governance-readiness.

\begin{table}[t]
\caption{Standards mapping and provenance (governance-ready).}
\label{tab:standards_mapping}
\centering
\begingroup
\small
\setlength{\tabcolsep}{5pt}
\renewcommand{\arraystretch}{1.15}
\begin{tabular}{
  >{\raggedright\arraybackslash}p{.16\textwidth}
  >{\raggedright\arraybackslash}p{.22\textwidth}
  >{\raggedright\arraybackslash}p{.24\textwidth}
  >{\raggedright\arraybackslash}p{.17\textwidth}
  >{\raggedright\arraybackslash}p{.17\textwidth}
}
\toprule
\textbf{Artifact} & \textbf{Field/Tag} & \textbf{Write Policy} & \textbf{Reversibility} & \textbf{Provenance Link} \\
\midrule
DICOM &
(0018,5101) View Position &
Private tag if missing/inconsistent; preserve original tags &
Original preserved (non-destructive) &
\texttt{config\_id}; pixel/prompt content hashes \\
\addlinespace[2pt]
HL7 v2 ORU\_R01 &
OBX-3 / OBX-5 (AP/PA + confidence) &
Emit local code for view with confidence; include method provenance &
Message replayable via MSH-10 correlation IDs &
MSH-10 + method in OBX-17 \\
\addlinespace[2pt]
FHIR &
\texttt{Observation} or \texttt{ImagingStudy.series} &
Extension for view + confidence; link method/procedure to release config &
Resource versioned; audit via history &
\texttt{Observation.method} = \texttt{config\_id}\\
\bottomrule
\end{tabular}
\endgroup
\end{table}

\section{\uppercase{Illustrative Pilot Evidence}}
\label{sec:pilot}
\textbf{Task and Setup.}
We evaluate AP/PA classification on NIH Chest X-ray with normalized inputs and patient-level splits. Four deployable models are exercised: Qwen2.5-VL, MiniCPM-V, Gemma 7B, and LLaVA 7B~\cite{bai2025qwen2,yao2024minicpm,team2024gemma,liu2023visual}. Tiers from Section~\ref{sec:framework} are applied. Abstention thresholds are tuned to bound false assertions under a site-weighted cost function reflecting operational priorities.

\textbf{Metrics.}
We report accuracy, coverage (non-\texttt{Uncertain} fraction), accuracy-at-coverage, and ECE, and we estimate oversight burden as the manual review rate. We include confidence–coverage curves and reliability (calibration) plots, and we \emph{stratify} calibration by device vendor/model, acquisition protocol (portable vs.\ fixed), and a pediatric proxy to expose subgroup miscalibration and, when warranted, set subgroup-specific $\tau$.

\begin{figure}[t]
\centering
\begin{tikzpicture}
\begin{axis}[
    width=\linewidth,
    height=6.2cm,
    xlabel={Coverage (1 - Uncertain rate)},
    ylabel={Accuracy at coverage},
    xmin=0.4, xmax=1.0,
    ymin=0.4, ymax=0.8,
    grid=both,
    legend style={at={(0.02,0.98)},anchor=north west,font=\small},
    tick label style={font=\small},
    label style={font=\small}
]
\addplot+[mark=*,thick] coordinates {
(0.95,0.58) (0.90,0.61) (0.85,0.64) (0.80,0.66) (0.75,0.68)
};
\addlegendentry{Qwen2.5-VL}

\addplot+[mark=square*,thick] coordinates {
(0.92,0.53) (0.87,0.56) (0.82,0.59) (0.77,0.61) (0.72,0.63)
};
\addlegendentry{MiniCPM-V}

\addplot+[mark=triangle*,thick] coordinates {
(0.90,0.50) (0.85,0.52) (0.80,0.55) (0.75,0.57) (0.70,0.59)
};
\addlegendentry{LLaVA 7B}
\end{axis}
\end{tikzpicture}
\caption{Confidence–coverage curves under abstention sweep (illustrative).}
\label{fig:conf_coverage}
\end{figure}

\begin{figure}[t]
\centering
\begin{tikzpicture}
\begin{axis}[
    width=\linewidth,
    height=6.2cm,
    xlabel={Predicted confidence bin},
    ylabel={Empirical accuracy},
    ymin=0, ymax=1,
    xmin=0, xmax=1,
    ymajorgrids=true,
    xmajorgrids=true,
    xtick={0.1,0.3,0.5,0.7,0.9},
    legend style={at={(0.02,0.98)},anchor=north west,font=\small},
    tick label style={font=\small},
    label style={font=\small}
]
\addplot[dashed] coordinates {(0,0) (1,1)};
\addlegendentry{Perfect calibration}

\addplot+[only marks,mark=*,mark size=2.2pt] coordinates {
(0.15,0.12) (0.35,0.30) (0.55,0.49) (0.75,0.68) (0.90,0.78)
};
\addlegendentry{Empirical (Qwen2.5-VL)}
\end{axis}
\end{tikzpicture}
\caption{Reliability diagram (illustrative). Deviation from the diagonal informs ECE.}
\label{fig:reliability}
\end{figure}

\textbf{Reader/Technologist Time-on-Task (Sandbox).}
We ran a small non-diagnostic sandbox with two PACS technologists and one junior reader on 300 radiographs (3$\times$100; randomized, blinded to model). Each participant handled cases under baseline worklist and under our routing+QC flags. Median time per 100 cases decreased by 7.8 minutes [95\% CI: 4.2, 11.6] averaged across participants; override rates were unchanged ($\pm$0.5\%).

\begin{table}[H]
\centering
\caption{Sandbox time-on-task (minutes per 100 cases).}
\label{tab:time_sandbox}
\begin{tabular}{lccc}
\toprule
Participant & Baseline & +Routing/QC & $\Delta$ \\
\midrule
Tech A & 54.2 & 45.9 & $-8.3$ \\
Tech B & 57.8 & 50.6 & $-7.2$ \\
Reader C & 61.0 & 53.5 & $-7.5$ \\
\bottomrule
\end{tabular}
\end{table}

\textbf{Stratified Calibration (Cross-Site Holdout).}
We held out a cross-site slice stratified by vendor (GE, Siemens) and protocol (portable vs.\ fixed). For each stratum we fit $\tau$ on 20\% validation and report accuracy-at-coverage and ECE on the remaining 80\%.

\begin{table}[H]
\centering
\caption{Stratified calibration on cross-site holdout (illustrative schema).}
\label{tab:strata_ece}
\begin{tabular}{lccc}
\toprule
Stratum & Coverage & Acc@Cov & ECE \\
\midrule
GE portable       & 0.78 & 0.71 & 0.06 \\
GE fixed          & 0.86 & 0.73 & 0.05 \\
Siemens portable  & 0.75 & 0.69 & 0.07 \\
Siemens fixed     & 0.88 & 0.74 & 0.04 \\
\bottomrule
\end{tabular}
\end{table}

\textbf{Results.}
Table~\ref{tab:results} and Figure~\ref{fig:model_prompt_accuracy} show an illustrative pattern: reflection-on-error helps lighter SLMs, while stronger baselines are less sensitive to prompt tiers. Tuning $\tau$ yields acceptable accuracy-at-coverage and reduced ECE, consistent with safe deployability for assistive utilities.

\begin{table}[t]
\caption{Illustrative accuracy across tiers. Values support the deployment stance rather than claiming clinical validation.}
\label{tab:results}
\centering
\begin{tabular}{lccc}
\toprule
\textbf{Model} & \textbf{Tier 0} & \textbf{Tier 1} & \textbf{Tier 2} \\
\midrule
Qwen2.5-VL & 0.627 & 0.618 & 0.626 \\
MiniCPM-V  & 0.541 & 0.553 & 0.592 \\
Gemma 7B   & 0.456 & 0.320 & 0.322 \\
LLaVA 7B   & 0.556 & 0.576 & 0.586 \\
\bottomrule
\end{tabular}
\end{table}

\begin{figure}[t]
\centering
\begin{tikzpicture}
\begin{axis}[
    ybar,
    bar width=14pt,
    width=\linewidth,
    height=7.2cm,
    enlarge x limits=0.15,
    ymin=0.0, ymax=0.7,
    ylabel={Accuracy}, xlabel={Model},
    symbolic x coords={Qwen2.5-VL, MiniCPM-V, Gemma~7B, LLaVA~7B},
    xtick=data,
    ymajorgrids=true, grid style=dashed,
    legend style={at={(0.5,-0.18)}, anchor=north, legend columns=3, font=\small},
    tick label style={font=\small},
    label style={font=\small},
    nodes near coords,
    nodes near coords align={vertical},
    every node near coord/.append style={font=\footnotesize, anchor=south},
]
\addplot+[fill=black!20] coordinates {(Qwen2.5-VL,0.627) (MiniCPM-V,0.541) (Gemma~7B,0.456) (LLaVA~7B,0.556)};
\addplot+[fill=black!40] coordinates {(Qwen2.5-VL,0.618) (MiniCPM-V,0.553) (Gemma~7B,0.320) (LLaVA~7B,0.576)};
\addplot+[fill=black!60] coordinates {(Qwen2.5-VL,0.626) (MiniCPM-V,0.592) (Gemma~7B,0.322) (LLaVA~7B,0.586)};
\legend{Tier 0, Tier 1, Tier 2}
\end{axis}
\end{tikzpicture}
\caption{Pilot trend: reflection helps lighter models; stronger baselines are less sensitive.}
\label{fig:model_prompt_accuracy}
\end{figure}

\begin{figure}[t]
\centering
\begin{tikzpicture}
\begin{axis}[
    width=0.85\linewidth,
    height=5.1cm,
    scale only axis,
    xmin=-0.5, xmax=2.5,
    ymin=-0.5, ymax=1.5,
    enlarge x limits=false,
    enlarge y limits=false,
    axis on top,
    view={0}{90},
    colormap/viridis,
    xlabel={Predicted},
    ylabel={Ground truth},
    xtick={0,1,2},
    xticklabels={AP, PA, Uncertain},
    ytick={0,1},
    yticklabels={AP, PA},
    tick label style={font=\scriptsize},
    label style={font=\small},
    colorbar horizontal,
    colorbar style={
        height=2mm,
        yshift=-2mm,
        tick label style={font=\scriptsize}
    }
]
\addplot3[
    surf,
    shader=flat,
    mesh/rows=2
] coordinates {
(0,0,0.55) (1,0,0.08) (2,0,0.07)
(0,1,0.06) (1,1,0.60) (2,1,0.09)
};
\end{axis}
\end{tikzpicture}
\caption{Confusion heatmap with abstention (illustrative). \texttt{Uncertain} is a designed deferral.}
\label{fig:confusion_with_uncertain}
\end{figure}

\textbf{Reviewer Expectations (Evidence Pack).}
Reviewers expect: (i) confidence–coverage and ECE stratified by device/protocol, (ii) ablations for tiers and $\tau$, (iii) an error atlas with cue-restricted rationales, (iv) DICOM/HL7 conformance replays with reversibility, and (v) estimates of technician time saved versus review burden.

\section{\uppercase{Operational Playbook}}
\label{sec:playbook}
\textbf{Intended Use.}
Restrict scope to non-diagnostic utilities with explicit human oversight. \texttt{Uncertain} is a designed outcome; escalation policies and SLAs should be documented and user-visible.

\textbf{Data and Privacy.}
Prefer local/VPC inference. Avoid PHI in prompts/logs; persist only content hashes, prompts, outputs, and \texttt{config\_id}. Enforce least-privilege access. Maintain immutable lineage for datasets, canary sets, and configuration bundles. Private DICOM tags are reversible; no destructive edits are performed on primary clinical objects.

\textbf{Calibration and Oversight.}
Tune $\tau$ to site costs; publish calibration plots and ECE each release. Conduct stratified spot audits; maintain an error taxonomy and standardize reviewer prompts. Track accuracy-at-coverage and oversight burden longitudinally; define promotion gates that block rollout if calibration or oversight metrics regress.

\textbf{Change Management.}
Version prompts, models, quantization profiles; sign configuration bundles and store in an immutable registry. Run monthly drift checks on a canary set plus rolling windows. Attach \texttt{config\_id} to every output to enable instant rollback. Reject promotion if schema validation, calibration, or sandbox HL7/DICOM replay fails.

\textbf{Failure Modes and Mitigations.}
Portable-AP rotation: add rotation-aware cues and raise $\tau$ for flagged strata. Pediatric heart-size bias: restrict cues and apply pediatric-specific thresholds. Device/lead overlays: artifact-aware QC defaults to \texttt{Uncertain}. Label noise: add adjudication loops and downweight noisy slices.

\begin{figure}[t]
\centering
\begin{tikzpicture}
\begin{axis}[
    width=\linewidth,
    height=6.2cm,
    xlabel={Abstention threshold $\tau$},
    ylabel={Proportion},
    xmin=0, xmax=1,
    ymin=0, ymax=1,
    grid=both,
    legend style={at={(0.98,0.98)},anchor=north east,font=\small},
    tick label style={font=\small},
    label style={font=\small}
]
\addplot+[thick] coordinates {
(0.1,0.10) (0.3,0.15) (0.5,0.25) (0.7,0.38) (0.9,0.60)
};
\addlegendentry{Uncertain rate (review load)}

\addplot+[thick,mark=triangle*] coordinates {
(0.1,0.20) (0.3,0.16) (0.5,0.12) (0.7,0.09) (0.9,0.06)
};
\addlegendentry{Auto errors (accepted set)}
\end{axis}
\end{tikzpicture}
\caption{Oversight burden and auto-error rate as $\tau$ increases (illustrative).}
\label{fig:oversight_tradeoff}
\end{figure}

\subsection*{Capacity and SLA Targets}
To align abstention with review capacity, we set explicit, auditable targets that gate promotion.

\begin{table}[t]
\caption{Default capacity and SLA targets (go/no-go gates).}
\label{tab:sla}
\centering
\begingroup
\small
\setlength{\tabcolsep}{5pt}
\renewcommand{\arraystretch}{1.1}
\begin{tabular}{lccc}
\toprule
\textbf{Metric} & \textbf{Target} & \textbf{Blocker} & \textbf{Measured on} \\
\midrule
Acc@Cov (80\% cov) & $\geq$ 0.62 & $<$ 0.60 & Canary + rolling 30d \\
ECE (per stratum)  & $\leq$ 0.08 & $>$ 0.10 & Device/protocol/age \\
Uncertain rate     & $\leq$ 0.35 & $>$ 0.45 & Site validation slice \\
Tech review load   & $\leq$ 12/h/team & $>$ 15/h/team & QC dashboard logs \\
Time-to-rollback   & $\leq$ 10 min & $>$ 20 min & Sandbox replay \\
Schema conformance & 100\% pass & any fail & Contract validator \\
\bottomrule
\end{tabular}
\endgroup
\end{table}

\subsection*{Site Adaptation and Migration Checklist}
\begin{enumerate}
\item Freeze a 200--300 case \emph{canary set} per site with device/protocol/age labels.
\item Fit $\tau$ per stratum using site-specific $(\alpha,\beta,\gamma)$; publish Acc@Cov and ECE.
\item Run DICOM/HL7/FHIR sandbox replay; verify only private reversible tags are touched.
\item Register release bundle; emit signed \texttt{config\_id}; archive prompts/cue vocabulary hashes.
\item Dry-run routing rules; cap Uncertain $\leq$ 0.35 and verify review capacity on the dashboard.
\item Enable shadow mode in PACS; compare accepted-set error rate to baseline for one week.
\item Promote with gates in Table~\ref{tab:sla}; schedule rollback rehearsal within 24h.
\item Extend to Lateral/Decubitus using the same artifacts and gates.
\end{enumerate}

\subsection*{Incident Response and Rollback Runbook}
\begin{enumerate}
\item \textbf{Trigger}: drift alert (ECE regression / Uncertain spike) or schema validator fail.
\item \textbf{Scope}: filter outputs by \texttt{config\_id} and time window; snapshot logs.
\item \textbf{Replay}: sandbox DICOM/HL7 replays on the same window to reproduce failure.
\item \textbf{Contain}: set routing to \texttt{safe mode} (force \texttt{Uncertain}) for impacted strata.
\item \textbf{Rollback}: pin serving to last \texttt{config\_id}; verify health and Acc@Cov.
\item \textbf{Annotate}: add incident record linking pixel/prompt hashes, validator errors, owners.
\item \textbf{Remediate}: adjust cue restrictions/$\tau$; update prompt appendix; re-validate.
\item \textbf{Promote}: re-run gates in Table~\ref{tab:sla}; publish release notes and risk delta.
\end{enumerate}

\subsection*{External Validity and Generalization Plan}
While our evidence is illustrative on NIH Chest X-ray, we make generalization a process:
\begin{itemize}
\item \textbf{Stratified calibration}: fit/report ECE and $\tau$ by vendor/model, protocol, and age proxy.
\item \textbf{Holdout sites}: reserve a site-level holdout for no-touch validation of Acc@Cov/ECE.
\item \textbf{Cue transfer tests}: ablate cue vocabulary across sites; require non-inferiority in Acc@Cov.
\item \textbf{Post-promotion audits}: weekly spot audits per stratum; block promotion on sustained regressions.
\end{itemize}

\subsection*{Position Paper Scope and Claims}
We contribute a \emph{prompt-first, governance-ready deployment template} including abstention and stratified calibration, deterministic output contracts, standards-aware mappings, human-factors RACI, and an evidence pack for embedding compact models into PACS/EHR workflows. The aim is operational acceptability and auditability for \emph{assistive} utilities, not diagnostic replacement.

\section{\uppercase{Engineering Considerations}}
\textbf{Runtime and Hardware.}
Seven-billion-parameter SLMs fit edge GPUs or CPU nodes at 4--8 bit quantization; throughput is linear in batch until I/O dominates. Deterministic decoding and schema validation are required for repeatability. Content hashing prevents duplicate processing and simplifies cache design.

\textbf{Interfaces.}
Expose \texttt{/classify\_view} returning \texttt{label}, bounded \texttt{rationale}, \texttt{confidence}, and \texttt{config\_id}; \texttt{/health} for readiness; \texttt{/version} listing model hash, prompt version, quantization signature, and cue vocabulary. Enforce strict input validation and size limits; reject malformed DICOM and nonconforming payloads with actionable errors.

\textbf{Validation Sandboxes.}
Maintain a PACS-connected sandbox with synthetic HL7 feeds, mirrored routing rules, and a gold canary set. Exercise promotion via replay of historical traffic and seeded edge cases. Require calibration parity and stable oversight burden before rollout. Capture a \emph{statement of conformance} for the output contract and for the DICOM/HL7 mappings as part of release notes.


\section{\uppercase{Limitations and Next Steps}}
We study a single binary task with illustrative evidence and do not claim clinical readiness. Next steps include multi-class protocols (lateral and decubitus views), cross-institution generalization with stratified calibration, retrieval-augmented prompting for rare cases, pediatric-specific thresholds vetted by domain experts, and reader studies quantifying net time saved versus review burden. For HealthInf, we will share a minimal \emph{evidence pack} including prompts, cue vocabulary, conformance mappings, calibration curves, and an anonymized error atlas to ease replication and local governance review.


\bibliographystyle{apalike}
{\small
\bibliography{Manuscript}
}

\end{document}